\documentclass[letterpaper, 10 pt, conference]{ieeeconf}
\IEEEoverridecommandlockouts    % This command is only needed if
\usepackage{graphics}           
\usepackage{times}              
\usepackage{amsmath}            
\usepackage{amssymb}            
\usepackage{graphicx}
\usepackage{algorithm}
\usepackage[noend]{algpseudocode}
\usepackage{booktabs}
\usepackage{color}
\usepackage[nocompress]{cite} % sorts citations automatically, e.g., "[1],[2],[3]" instead of "[2],[3],[1]".
\definecolor{instructioncolor}{rgb}{.5,.5,.5}
\usepackage{multirow} 
\usepackage{url}

\usepackage[font=small]{caption}
\usepackage{svg}
\def\eqref#1{Eq.~(\ref{#1})}

\makeatletter
\usepackage{xspace}
\DeclareRobustCommand\onedot{\futurelet\@let@token\@onedot}
\def\@onedot{\ifx\@let@token.\else.\null\fi\xspace}

\def\etal{{et al}\onedot}
\makeatother

\usepackage{array}
\newcolumntype{L}[1]{>{\raggedright\let\newline\\\arraybackslash\hspace{0pt}}m{#1}}
\newcolumntype{C}[1]{>{\centering\let\newline\\\arraybackslash\hspace{0pt}}m{#1}}
\newcolumntype{R}[1]{>{\raggedleft\let\newline\\\arraybackslash\hspace{0pt}}m{#1}}

\title{\LARGE \bf Semantic-Aware Particle Filter for Reliable Vineyard Robot Localisation}

\author{Rajitha de Silva$^{1}$ \and Jonathan Cox$^{1}$ \and James R. Heselden$^{1}$ \and Marija Popovi\'{c}$^{2}$ \and Cesar Cadena$^{3}$ \and Riccardo Polvara$^{1}$% <-this % stops a space
  \thanks{$^{1}$Rajitha de Silva, Jonathan Cox, James R. Heselden and Riccardo Polvara are with Lincoln Centre for Autonomous Systems (L-CAS), University of Lincoln, UK. $^{2}$Marija Popovi\'{c} is with MAVLab, TU Delft, Netherlands. $^{3}$Cesar Cadena is with Robotics Systems Lab, ETH Zurich, Switzerland. (corresponding author: Rajitha de Silva {\tt\small $^{1}$odesilva@lincoln.ac.uk})}%
  \thanks{This work was supported by Engineering and Physical Sciences Research Council (EPSRC), UK Project: ``GAIA: Ground-Aerial maps Integration for increased Autonomy outdoors" (EPSRC Reference: EP/Y003438/1).}% <-this % stops a space
}

\begin{document}
\maketitle
\thispagestyle{empty}
\pagestyle{empty}

%%%%%%%%%%%%%%%%%%%%%%%%%%%%%%%%%%%%%%%%%%%%%%%%%%%%%%%%%%%%%%%%%%%%%%%%%%%%%%%%
\begin{abstract}
  %
  %% WHY 
  % Use 1-2 not too long sentences, which clearly answer the WHY question: 
  % Why is this relevant, why should I care? Motivate why the stuff that 
  % you enable is relevant (not neccessarily equal to the technique)

  %% WHICH PROBLEM 
  % One sentence that explain the problem the paper addresses/ivestigates
  % Start with: In this paper, we address the problems of \dots
Accurate localisation is critical for mobile robots in structured outdoor environments, yet LiDAR-based methods often fail in vineyards due to repetitive row geometry and perceptual aliasing. We propose a semantic particle filter that incorporates stable object-level detections, specifically vine trunks and support poles into the likelihood estimation process. Detected landmarks are projected into a bird’s eye view and fused with LiDAR scans to generate semantic observations. A key innovation is the use of semantic walls, which connect adjacent landmarks into pseudo-structural constraints that mitigate row aliasing. To maintain global consistency in headland regions where semantics are sparse, we introduce a noisy GPS prior that adaptively supports the filter. Experiments in a real vineyard demonstrate that our approach maintains localisation within the correct row, recovers from deviations where AMCL fails, and outperforms vision-based SLAM methods such as RTAB-Map.

  %% HOW & WHAT
  % Around 3 sentences that explain how to approach the problem in general and answers:
  % How to solve the problem in general? (1/2 - 1 sentence)
  % What makes our approach special? What are we actually doing? What is new?
  
  %% IMPLEMENTATION, EVALUATION, WHAT FOLLOWS
  % 1-2 sentences what the experiments show and potentiall what follows from
  % your great work for the research community or the rest of the world ;-)

\end{abstract}

%%%%%%%%%%%%%%%%%%%%%%%%%%%%%%%%%%%%%%%%%%%%%%%%%%%%%%%%%%%%%%%%%%%%%%%%%%%%%%%%
\section{Introduction}
\label{sec:intro}

%%%%%%%%%%%%%%%%%%%
%% WHY: 
% First, answer the WHY question: Why is that relevant? Why should I be
% motivated to read the paper? Why should I care? (1 paragraph, 2-5 sentences)

Accurate localisation is a critical component of mobile robot navigation in outdoor environments~\cite{aguiar2022localization}. Among the various approaches, LiDAR-based localisation remains widely adopted due to its reliable and precise perception of geometric structure. However, these methods rely solely on scene geometry, which can be problematic in outdoor agricultural settings like vineyards, where repetitive and ambiguous structures are common~\cite{nehme2021lidar}. In such environments, incorporating semantic information complements the geometric structure offering a promising alternative to enhance localisation performance~\cite{de2025keypoint}.

%%%%%%%%%%%%%%%%%%%
%% WHICH PROBLEM
% Second, explain WHICH problem you are solving/address to solve.

In this paper, we tackle the challenge of semantic ambiguity in geometry-based localisation within vineyard environments. The repetitive structure of vineyard rows often induces perceptual aliasing in LiDAR range data, resulting in localisation drift and errors. To overcome this limitation, we exploit semantically meaningful landmarks, specifically vine trunks and support poles whose distinctive spatial distributions provide stronger discriminative cues. Our approach detects these semantic objects and estimates their relative positions from RGB-D imagery, which are then projected onto the LiDAR frame. This enables a semantic-LiDAR particle filter that offers a robust alternative to conventional localisation methods.

\begin{figure}[t]
  \centering
  \includegraphics[width=0.99\linewidth]{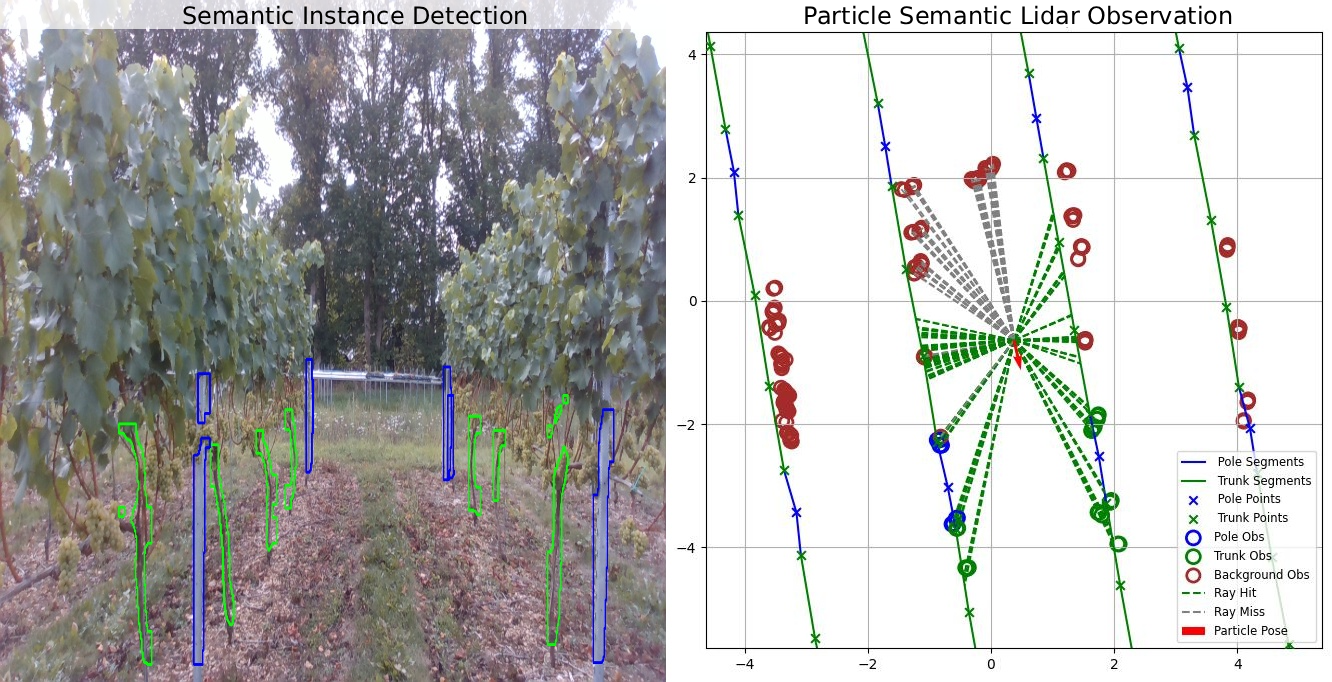}
  \caption{Vineyard Semantics detection (left) projected onto the landmark map (right) for likelihood estimate calculation. Green: Vine trunks, Blue: Support poles.}
  \label{fig:motivation}
\end{figure}

%%%%%%%%%%%%%%%%%%%
%% HOW & WHAT
% Third, explain briefly how one can address the problem in general and mention 
% briefly what others/we before have done. Prepare the reader for your contribution 
% that comes in the next section (and not here!).
Traditional particle filters, such as Adaptive Monte Carlo Localisation (AMCL)~\cite{amcl}, estimate a robot’s pose by evaluating the geometric consistency between sensor observations and a known map, an approach that has proven highly effective in structured indoor and urban settings where distinctive geometric features are abundant. Vineyards, however, present a markedly different challenge: their long, repetitive rows induce strong perceptual aliasing, while unstable elements such as foliage and grape clusters provide little reliability for long-term localisation~\cite{papadimitriou2022loop}. We contend that robust localisation in such environments requires moving beyond raw geometry and explicitly exploiting semantics. Our key insight is that vine trunks and support poles serve as stable, distinctive landmarks whose consistent spatial distribution across rows can disambiguate pose estimates. Moreover, we introduce the concept of semantic walls, where the space between consecutive landmarks is modelled as a pseudo-structural boundary. This transforms sparse semantic detections into continuous row-level constraints, creating a representation that is far more robust to vineyard aliasing and seasonal variation. Together, these ideas lay the foundation for a semantic-LiDAR particle filter that redefines localisation in repetitive agricultural environments, as illustrated in Fig.~\ref{fig:motivation}.

% Link to figure somewhere
% See \figref{fig:motivation} for an example.

%%%%%%%%%%%%%%%%%%%
%% MAIN CONTRIBUTION & WHAT FOLLOWS FROM THAT
% Explain your contribution in one paragraph. This is a very important paragraph. 
% Always start that paragraph with: ``The main contribution of this paper is''
% and be SUPER-EXPLICIT for what you claim contribution (and maybe novelty) for.

The main contributions of this paper are threefold:
(i) the design of a semantic particle filter that integrates object-level detections of stable vineyard landmarks (trunks and support poles) with a 2D semantic map, enabling robust localisation in highly repetitive environments;
(ii) the introduction of the semantic walls concept, which augments sparse landmark detections by modelling pseudo-structural boundaries between adjacent landmarks, thereby strengthening row-level constraints and mitigating perceptual aliasing; and
(iii) a systematic evaluation against established baselines, demonstrating the trade-offs in accuracy, robustness, and sensor requirements.

%%%%%%%%%%%%%%%%%%%
%% OUR KEY CLAIMS (can be merged with the main contribution above if desired)
% Explicitly(!) state your claims in one (short) paragraph and make
% sure you pick them up again in the experiments and support every claim.

%In sum, we make three key claims:
%Our approach is able to
%
%(i) \dots; our approach can filter point belong to fast and slow dynamic achieving real time performance.  
%
%(ii) \dots; can improve the localisation performance when localising with previous outdated map robustly track robot pose in highly visually appearance changing environment. 
%
%(iii) \dots.
%
%These claims are backed up by the paper and our experimental evaluation.

%%%%%%%%%%%%%%%%%%%%%%%%%%%%%%%%%%%%%%%%%%%%%%%%%%%%%%%%%%%%%%%%%%%%%%%%%%%%%%%%
\section{Related Work}
\label{sec:related}

% Discuss the main related work and cite around key papers. The number
% of cites depend on the page limit. Never below 15, 20-30 is often fine
% for conference papers.
% The related work section should be approx. 1 column long, assuming 
% a 6-page paper, up to one page for 8-10 page papers.  You can structure 
% the  section in paragraphs, grouping the  papers, and describing the key 
% approaches with 1-2 sentences. Avoid dull enumerations of papers. If 
% applicable, describe the key difference to your approach at the end 
% of each paragraph briefly. Avoid adding subsections, al least for a 
% conference paper.
%% BRIEFLY SUMMARIsE OWN CONTRIBUTION AT THE END

\textbf{Robot localisation in vineyards} presents several challenges due to the environment’s repetitive row structure, dynamic seasonal changes, and unfavourable terrain conditions~\cite{kokas2024multicamera, agriculture11100954}. Many of these challenges are sensor-specific. For example, visual sensors such as LiDAR and cameras are prone to perceptual aliasing in repetitive environments and are sensitive to variable lighting and seasonal appearance changes~\cite{hroob2021benchmark, shi2025vision}. Non-visual sensors, including GPS and ultrasonic sensors, can be affected by canopy occlusion, signal multipath or reflection errors, and limited sensing range~\cite{agronomy11020287, costley2020landmark}. Additionally, environmental factors such as uneven terrain and wheel slippage can degrade localisation accuracy regardless of the sensor modality~\cite{rakun2022sensor}. Various localisation approaches have been applied in vineyard environments, including LiDAR-based particle filters such as AMCL~\cite{astolfi2018vineyard, s22239095}, visual SLAM methods~\cite{kokas2024multicamera, automation4040018}, and GPS-based systems~\cite{rakun2022sensor, 9249176}. Some systems avoid explicit localisation altogether by relying on reactive navigation strategies, typically alternating between row-following and row-switching behaviours~\cite{10341261}. However, robust localisation remains essential for enabling long-term autonomy and supporting digital twin applications in vineyard management~\cite{polvara2024bacchus, kunze2018artificial}. Among the various approaches, particle filters continue to be a useful and adaptable method for robot localisation in vineyard environments~\cite{aguiar2022localization}.

\textbf{Particle filters in agricultural environments} are widely used due to their ability to model complex, non-Gaussian sensor behaviors arising from factors such as wheel slippage and uneven terrain~\cite{cerrato2020gps}. In vineyard settings, the absence of smooth, consistent surfaces and the presence of dense foliage often result in imperfect maps. A robot attempting to localise within such maps using deterministic methods may fail due to geometric inconsistencies introduced by temporal changes, such as seasonal growth. The probabilistic nature of particle filters allows them to tolerate map-sensor mismatches, making them well-suited for environments with dynamic vegetation and appearance variations~\cite{blok2019robot}. However, applying particle filters at large scales in vineyard environments can be computationally expensive and inefficient. The need to maintain a high number of particles to cover a large area, combined with the repetitive nature of vine rows, can lead to incorrect pose estimates. To address this, introducing a noisy GPS prior can help constrain the initial particle distribution, preserving the benefits of probabilistic localisation while improving efficiency in large-scale deployments~\cite{oh2004map}.

\textbf{Use of semantic information} can significantly enhance the accuracy of particle filter-based localisation methods~\cite{huang2024semantics}, particularly in geometrically repetitive environments like vineyards. While geometric perception techniques such as LiDAR rely solely on depth measurements relative to the robot, semantic object detection can identify meaningful landmarks such as vine trunks or support poles while ignoring variable elements like foliage that are prone to seasonal changes. Papadimitriou \etal~\cite{papadimitriou2022loop} proposed a graph-SLAM approach that utilises depth information from semantically significant landmarks, specifically vine trunks, and demonstrated superior performance compared to non-semantic baselines. Similarly, De Silva \etal~\cite{de2025keypoint} introduced a keypoint-semantic integration method that improves visual feature matching in vineyards, highlighting the potential of semantic cues in enhancing localisation robustness. These works suggest that leveraging semantically relevant landmarks in place of purely geometric depth sensing can lead to more reliable localisation in vineyard environments.

%%%%%%%%%%%%%%%%%%%%%%%%%%%%%%%%%%%%%%%%%%%%%%%%%%%%%%%%%%%%%%%%%%%%%%%%%%%%%%%%
\section{Semantic Particle Filter in Vineyards}
\label{sec:main}
\begin{figure}[t]
  \centering
  \includegraphics[width=0.99\linewidth]{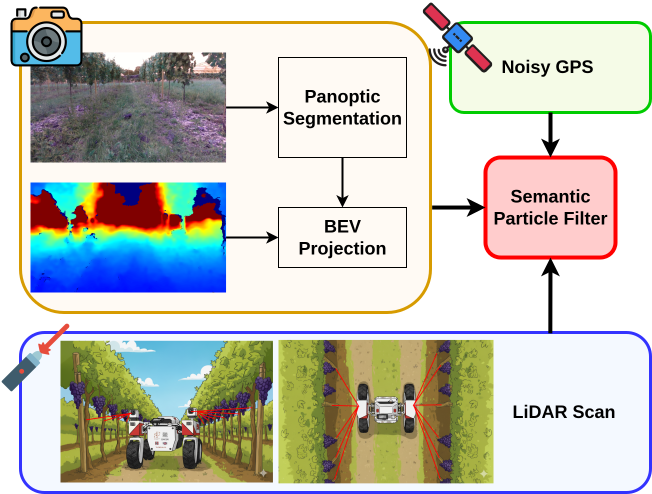}
  \caption{Overview of the Semantic Particle Filter. Instance masks obtained from panoptic segmentation are combined with depth data to generate pole-like BEV projections of vineyard landmarks. These projections serve as semantic observations within a particle filter framework, augmented with a noisy GPS prior to improve localisation robustness. LiDAR scans can be optionally incorporated to further enhance the quality of observations.
}
  \label{fig:overall}
\end{figure}

We propose a localisation framework that leverages long-term stable vineyard landmarks specifically vine trunks and support poles detected using panoptic segmentation. Following the approach of~\cite{7759339}, the detected landmarks are projected into a bird’s-eye view (BEV) representation. Each landmark in the BEV is then converted into a semantic mask, which is overlaid onto the LiDAR scan to distinguish semantic observations from background observations. The resulting semantic-LiDAR scan serves as the observation for the particle filter, enabling robust localisation in repetitive vineyard rows. To improve efficiency and constrain the global search space, we further introduce a dynamic weighting scheme that integrates a noisy GPS prior. In regions with limited semantic cues, such as headlands where the robot transitions between rows, the likelihood estimation is biased more heavily toward the GPS prior to maintain global consistency. Conversely, when the robot enters a row rich in semantic landmarks, the influence of the GPS prior is reduced, allowing localisation to rely primarily on semantic-LiDAR observations. The overall pipeline is illustrated in Fig.~\ref{fig:overall}.

%% Describe your approach. It is okay to divide the main section
%%  into a few subsections (e.g., 2-4 subsections).
\subsection{Panoptic Segmentation in Vineyards}
The semanticBLT dataset was collected in Ktima Gerovassiliou vineyard in Greece. To capture long-term stable landmarks, we trained a YOLOv9-based semantic instance segmentation model~\cite{yolov9} on the SemanticBLT dataset~\cite{semanticblt_dataset} to generate instance masks for two vineyard-relevant classes: poles and trunks. The model achieved higher detection performance for trunks reporting average precision (AP) of 0.691 compared to poles (AP = 0.573), reflecting the greater visual distinctiveness and stability of trunks, while poles proved more challenging due to their smaller size and visual similarity to background clutter. The panoptic segmentation module is designed to be modular, enabling the YOLOv9 model to be replaced with any other segmentation architecture without impacting the overall localisation pipeline.

\subsection{Bird's Eye View Projection}
For each image frame, detected landmarks are projected from the 2D image plane into a BEV map centred on the robot. Let a detected landmark $i$ be defined by its segmentation mask $M_i$ in the image plane and its class $c_i \in \{\text{pole, trunk}\}$. First, we determine the landmark's distance from the camera. Given a depth image $D$, the set of depth values corresponding to the mask $M_i$ is extracted. The landmark's distance, $d_i$, is defined as the minimum valid depth value within this set:
\begin{equation}
    d_i = \min \{D(p) \mid p \in M_i, D(p) > 0\}
\end{equation}

where $p=(u,v)$ is a pixel coordinate within the mask $M_i$. This provides an accurate estimate of the object's range by measuring its closest point. This approach also provides robustness against segmentation inaccuracies, as parts of the mask that may erroneously include the distant background will be ignored by the minimum operation.

Given the depth value \(d_i\) at \(p\), the 3D point \(\mathbf{p}_\text{cam} = [x, y, z]^\top\) in the camera coordinate frame is computed using the pinhole camera model as:

\begin{equation}
\begin{bmatrix}
x \\
y \\
z
\end{bmatrix}
=
d_i \cdot
\begin{bmatrix}
\frac{v - c_x}{f_x} \\
\frac{u - c_y}{f_y} \\
1
\end{bmatrix}
\end{equation}

Here, \((c_x, c_y)\) are the principal point offsets and \((f_x, f_y)\) are the focal lengths of the camera, all derived from the intrinsic calibration matrix. To obtain the BEV-relative position of the object in the robot frame, we assume a forward-facing camera configuration where the \(z\)-axis points forward and the \(x\)-axis points to the left of the robot. Accordingly, the BEV coordinates \((x_{\text{BEV}}, z_{\text{BEV}})\) are defined as: (\(-x, z)\). 

Following this projection, a set of BEV semantic masks is generated by inflating each detected landmark into a circular region of radius $\mathcal{R}_{sem}$, set to 0.5m. Points in the LiDAR scan that fall within a mask are assigned the semantic class of that landmark, while points outside all masks are labelled as background. This process yields a semantic scan, where each point is associated with a class label $c_i$. The semantic scan constitutes the final observation of the environment used for localisation.

\subsection{Semantic Vineyard Map}
\label{sec:semantic_map}

In traditional localisation methods such as AMCL, the global map is represented as an occupancy grid constructed from LiDAR scans acquired while the robot is manually driven through the environment. In contrast, we propose a pseudo-structural semantic map derived from a pre-existing Real-Time Kinematic GPS (RTK-GPS) survey of the vineyard. This map consists of georeferenced point locations of stable landmarks, specifically vine trunks and support poles. Such landmarks can be obtained through different means: for example, by direct surveying using accurate GPS measurements, by semi-autonomous mapping where semantic object detections from RGB-D imagery are used to infer landmark positions relative to the robot’s pose before projecting them onto a global frame, or by aerial perception methods that detect and map landmarks from overhead imagery. However, as these landmarks are relatively sparse compared to dense occupancy maps, we introduce "semantic walls" by connecting adjacent landmarks into pseudo-structural constraints. This provides continuous row-level cues that strengthen localisation and reduce ambiguity in repetitive vineyard environments.

\begin{figure}[t]
  \centering
  \includegraphics[width=0.99\linewidth]{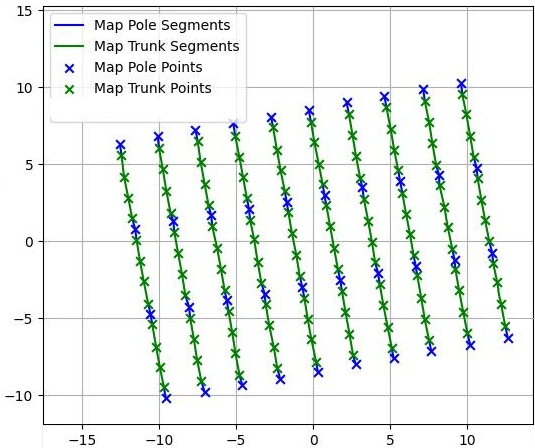}
  \caption{Illustration of the semantic wall map: landmarks (trunks and poles) are grouped and ordered by row, then connected into continuous line segments that serve as semantic obstacles for localisation..}
  \label{fig:wallmap}
\end{figure}

First, individual landmarks are grouped according to their designated row. Within each row, the landmarks are then ordered by spatial position to establish a sequential progression from one end to the other, producing an ordered list in which each landmark retains its semantic class. These sequential landmarks are subsequently connected to form line segments, creating a continuous semantic wall for each row, as illustrated in Fig.~\ref{fig:wallmap}. The semantic class of each segment is inherited from the landmarks it connects; for example, a segment between two trunks is labelled as \textit{trunk}, while a segment between a trunk and a pole is labelled as \textit{pole}. For the localisation algorithm, each line segment is treated as a solid obstacle, providing a practical and efficient abstraction of the vineyard’s highly structured and repetitive geometry. This constrained pseudo-structural representation compels the particle filter to estimate the robot’s pose using reliable local observations, while reducing row-level aliasing that arises from noisy or ambiguous measurements of adjacent rows.

\subsection{Particle Filter Likelihood Estimation}
Following standard particle filter approach, a set of particles $\mathbf{X}$ are initiated where every particle represents a hypothesis of the robot's pose. A LiDAR-semantic likelihood score and a GPS likelihood score is calculated for each particle.

\subsubsection{LiDAR-Semantic Likelihood}
\label{sec:sem_like}
For a given particle $\mathbf{x} \in \mathbf{X}$, we evaluate the set of observations $\mathcal{O} = \{\mathbf{o}_1, \ldots, \mathbf{o}_N\}$. Each observation $\mathbf{o}_i$ is defined by class, range and bearing [$c_i, r_i, \beta_i$].

For each observation, a ray is cast from the particle's position into the world. The ray's direction is determined by a local bearing $\beta_i$ relative to the particle while its extent is limited to a distance of 5~m. We then find the first intersection of this ray with the landmark map $\mathcal{M}$, which yields the distance to the map feature, $r_i^{\mathcal{M}}$, and the class of that feature, $c_i^{\mathcal{M}}$. The log-likelihood for this single observation, $\log p(\mathbf{o}_i | \mathbf{x}, \mathcal{M})$, is determined by one of three cases:

\begin{enumerate}
    \item Correct Hit: The ray intersects a map landmark of the same class ($c_i^{\mathcal{M}} = c_i$). The likelihood is a Gaussian function of the range error, $\Delta r_i = \left| r_i - r_i^{\mathcal{M}}\right|$.
    \item Incorrect Hit: The ray intersects a landmark of a different class ($c_i^{\mathcal{M}} \neq c_i$). A fixed penalty $\lambda_h$ is applied.
    \item Miss: The ray does not intersect any landmark. A fixed penalty $\lambda_m$ is applied.
\end{enumerate}

This is formally expressed as:

\begin{equation}
\log p(\mathbf{o}_i | \mathbf{x}, \mathcal{M}) = -\frac{(\lambda)^2}{2\sigma_{\text{obs}}^2}
\begin{cases}
\lambda=\Delta r_i & \text{if Correct Hit} \\
\lambda=\lambda_h & \text{if Incorrect Hit} \\
\lambda=\lambda_m & \text{if Miss}
\end{cases}
\label{eq:cases}
\end{equation}

where $\sigma_{\text{obs}}$ is the standard deviation for range error, and $\lambda_h$ and $\lambda_m$ are penalty constants. For the observations in the semantic landmark classes follow all the three cases of Equation~\ref{eq:cases} while the background class only follow hit and miss cases.

The total semantic log-likelihood for particle $\mathbf{x}$ is the weighted average over all $N$ observations:

\begin{equation}
\log p_{\text{obs}} = \frac{1}{N} \sum_{i=1}^{N} \omega_{c_i} \log p(\mathbf{o}_i | \mathbf{x}, \mathcal{M})
\end{equation}

where $\omega_{c_i}$ is a predefined weight for the observation's class. We assume higher importance weight for poles compared to trunks, based on the assumption that the artificial landmarks (poles) are more robust than the natural landmarks (trunks). The weight for the background class is the lowest of all the three. 

\subsubsection{GPS Likelihood}
The noisy GPS reading provides a spatial prior for global localisation. Given a GPS measurement $\mathbf{x}_{\text{GPS}} = [x_g, y_g]^\top$ with standard deviation $\sigma_{\text{GPS}}$, the GPS log-likelihood is:

\begin{equation}
\log p_{\text{GPS}} = -\frac{1}{2\sigma_{\text{GPS}}^2} \left( (x_i - x_g)^2 + (y_i - y_g)^2 \right)
\end{equation}

The total log-likelihood $\log \mathcal{L}_\mathbf{x}$ for particle $\mathbf{x}$ is a weighted sum of normalised the semantic and GPS likelihood components:

\begin{equation}
\log \mathcal{L}_\mathbf{x} = (1-\alpha) \log p_{\text{obs}} + \alpha \log p_{\text{GPS}}
\label{eq:lx}
\end{equation}
where $\alpha \in [0.05, 9.95]$ is a weighting factor balancing the two information sources. The $\alpha$ value is dynamically adjusted for each particle based on Equation~\ref{eq:alpha}:

\begin{equation}
\alpha = \frac{1}{1 + (N_{sem}/K)}
\label{eq:alpha}
\end{equation}

where $N_{sem}$ is the number of semantic observations and $K$ is a scaling constant. In our experiments $K$ was set to 4 such that, at least 4 semantic observations are detected when the robot is within a vineyard row. The final particle weights are then computed and normalised using the softmax function:

\begin{equation}
w_\mathbf{x} = \frac{\exp(\log \mathcal{L}_\mathbf{x})}{\sum_{k=1}^{N_p} \exp(\log \mathcal{L}_k)}
\end{equation}
where $N_p$ is the total number of particles.
%%%%%%%%%%%%%%%%%%%%%%%%%%%%%%%%%%%%%%%%%%%%%%%%%%%%%%%%%%%%%%%%%%%%%%%%%%%%%%%%
\section{Experimental Evaluation}
\label{sec:exp}

%% Repeat the main focus/objective with one single(!) sentence starting with:
%% Explain the reader that the experiments with support all claims
%% (same list as in the intro!) starting the paragraph with:}

The main focus of this work is to develop a semantic particle filter that uses semantic landmark detections for robot localisation in vineyard environments. We present our experiments to demonstrate that incorporating semantics to a particle filter improves the localisation accuracy. 

\subsection{Experimental Setup}
We used a Thorvald robot platform from Saga Robotics, equipped with a front-mounted Intel RealSense D435i camera, a 2D LiDAR, and a RTK-GPS, to conduct our experiments. The robot was driven through a small test vineyard consisting of 10 rows, during which RGB-D images, LiDAR scans, and RTK-GPS coordinates were recorded. This data was used to evaluate our semantic particle filter, with AMCL, RTAB-Map~\cite{rtabmap} and noisy GPS serving as the baselines. To simulate standard GPS from RTK ground-truth, we added random noise corresponding to a 2.0 m circular error probable (CEP). This produces trajectories that reflect the typical accuracy and uncertainty of consumer-grade GPS measurements.

%% If needed (and only then!) say also a few words about the experimental
%% setup, the datasets, and used parameters. You can use a separate subsection if you
%% want to put the focus on that but often that is not needed.}

%% Note 1: It MUST be always crystal clear (a) WHY an experiment is there
%% (e.g., to support a claim, to show that the approaches useful for real-word
%% systems, to show the performance, or to provide a baseline comparison), (b)
%% WHAT it wants to show (which claim/property exactly), and (c) HOW it aims at 
%% showing this. This is ESSENTIAL for a good evaluation. Think about when BEFORE
%% designing an experiment.  IMPORTANT: Every experiment MUST start with something 
%% like:  The next experiment is presented to show \dots and thus for supporting our 
%% first claim.

%% Note 2: Start with the most important/impressive experiment first. Make
%% his a key story of the paper. Keep the order of the claims, i.e., re-order
%% claims in the intro/before if needed. 

%%%%%%%%%%%%%%%%%%%%%%%%
\subsection{Results and Discussion}

\begin{figure*}[t]
  \centering
  \includegraphics[width=0.99\linewidth]{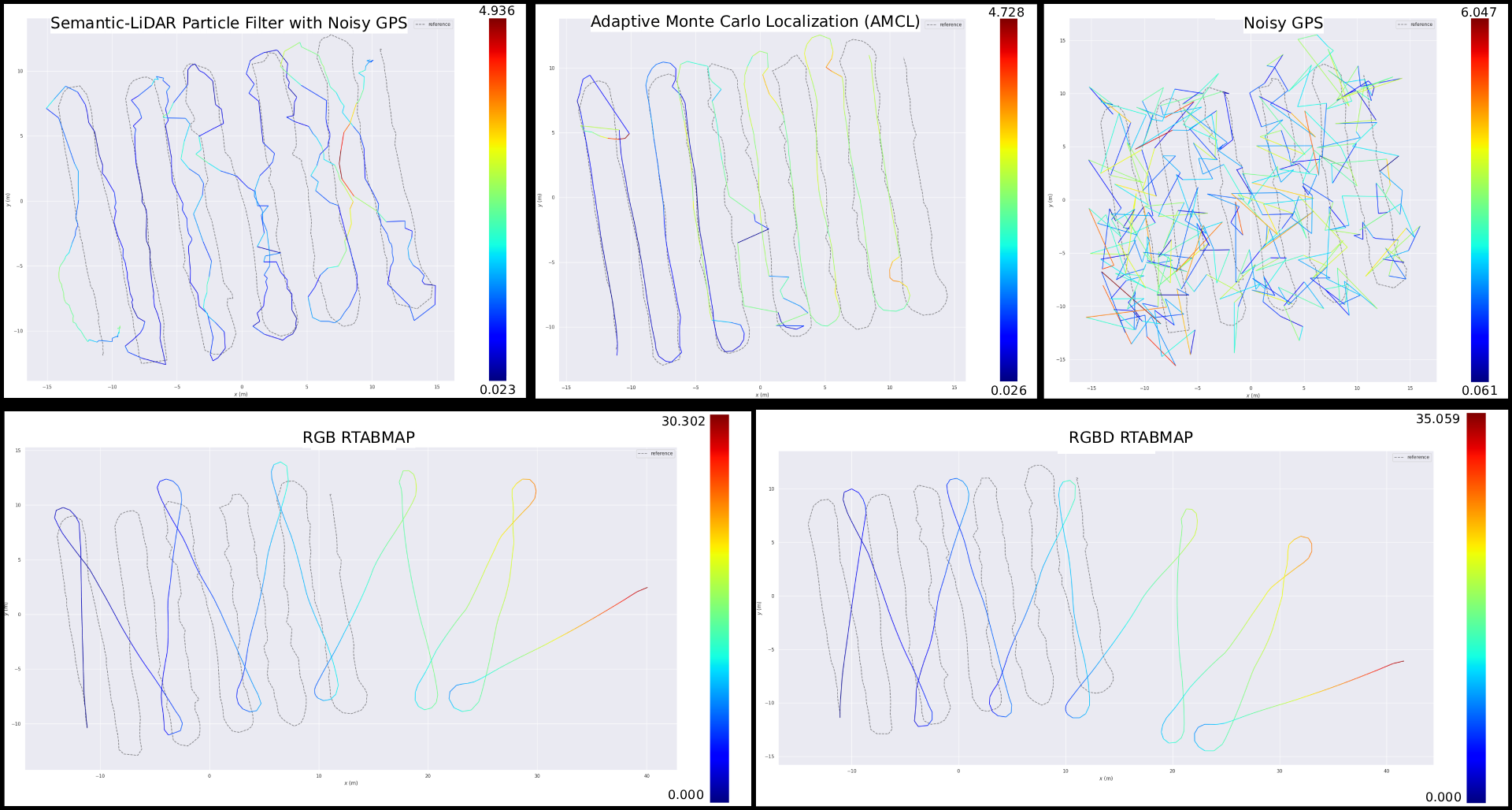}
  \caption{Absolute Pose Error (APE) trajectories generated using the EVO tool for different localisation algorithms. The comparison includes Semantic-LiDAR Particle Filter with Noisy GPS (Ours), Adaptive Monte Carlo Localisation (AMCL), Noisy GPS, RGB RTAB-Map, and RGB-D RTAB-Map. The dashed black line represents the ground-truth reference trajectory, while the color-coded paths show the deviation of each method.}
  \label{fig:plots}
\end{figure*}

%% First experiment - most impressive, important or the most important
%% claim supporting experiments comes first.
The experiment evaluates the performance of our approach, and its outcomes support the claim that incorporating semantic information improves localisation accuracy in perceptually aliased environments. We compare the estimated trajectories of the semantic particle filter against AMCL, RTAB-Map and noisy GPS baselines. Each method is compared against ground truth positions derived from the original RTK-GPS data. The results, summarised in Table~\ref{tab:evo}, show that the semantic particle filter outperforms all the baselines in both APE metric computed using the \texttt{evo}~\cite{evo} evaluation framework. However, the RPE of the semantic particle filter reports higher error owing to the noisy GPS biasing during the row switching stage of the navigation. The best RPE error was seen in RGB RTAB-Map. Our method outperforms the AMCL demonstrating a 33\% improvement, indicating more accurate and stable global localisation along the vineyard rows where geometry based particle filters struggles due to repetitive structure. It is worth noting that the default particle count range in AMCL is set between 100 and 5000, whereas our method achieved higher accuracy with a significantly smaller range of 80 to 500 particles. This highlights not only the superior accuracy of the proposed approach but also its computational efficiency. The proposed semantic particle filter achieves a 43\% reduction in APE compared to the noisy GPS baseline, while it outperforms the RGB and RGB-D RTAB-Map methods by 88\% and 90\%, respectively.

\begin{table}[t]
  \centering
  \caption{Comparison of relative pose error (RPE) and absolute pose error (APE) on semantic particle filter (SPF) and depth based particle filters (DPF). NGPS: Noisy GPS.}
  % Change the widths (in cm) to suit your layout needs
  \begin{tabular}{ccc}
    \toprule
    \multirow{2}{*}{\textbf{Method}} & \multicolumn{2}{c}{\textbf{Mean (Standard Deviation) [Unit: m]}}\\
    \cmidrule(lr){2-3}
    & APE & RPE \\
    \midrule
    %\textbf{SPF} + NGPS & \textbf{83.09(45.64)} & 44.92(36.29)\\
    \textbf{SPF + LiDAR + NGPS} & \textbf{1.20 (0.83)} & 0.94 (0.73)\\
    %DPF + NGPS & 261.15(209.36) & \textbf{39.27(40.10)}\\
    AMCL & 1.79 (1.09) & 0.23 (0.43)\\
    NGPS Only & 2.11 (1.10) & 3.03 (1.59)\\
    RGB RTAB-Map & 10.20 (6.62) & \textbf{0.16 (0.15)}\\
    RGBD RTAB-Map & 11.96 (4.23) & 0.70 (4.23)\\ 
    \bottomrule
  \end{tabular}
  \label{tab:evo}
\end{table}

As shown in Fig.~\ref{fig:plots}, AMCL produces smooth and consistent trajectories compared to the other methods. However, its principal limitation is the tendency to localise in the wrong row due to geometric aliasing, from which it is unable to recover once drift occurs. In contrast, our method consistently maintains localisation within the correct row and can also recover from deviations, owing to the dynamic GPS weighting incorporated into the particle filter. Both approaches are susceptible to occasional jumps in localisation, particularly along the outermost rows. The noisy GPS trajectory, as expected, remains bounded within its specified accuracy radius. By comparison, RTAB-Map exhibits the largest deviations, often diverging significantly from the vineyard structure due to overshoot during row switching and the effects of geometric aliasing.

In summary, our evaluation suggests that our method provides competitive localisation performance compared to traditional visual and LiDAR SLAM approaches, particularly in environments with repetitive geometric structure and appearance variability. By leveraging semantic landmarks and integrating a noisy GPS prior, the proposed semantic particle filter achieves more robust and accurate pose estimates under challenging vineyard conditions. These results highlight the potential of semantics-aware localisation for improving autonomy in structured agricultural environments.

\subsection{Limitations}
While the proposed semantic particle filter demonstrates strong performance in vineyard environments, several limitations remain. First, the approach relies on a surveyed semantic landmark map, which can be costly to generate and maintain at scale. This could be alleviated by integrating autonomous object detection and tracking methods to construct maps on demand. Second, the method is sensitive to missing or misclassified landmarks; significant occlusion or removal may reduce localisation accuracy. Nevertheless, the modular design of our system allows future advances in panoptic segmentation to be readily incorporated, mitigating this limitation. Finally, although our evaluation has been restricted to single-season vineyard trials, we expect the method to generalise across seasons. This expectation is supported by the fact that the panoptic segmentation model, which is the principal bottleneck for cross-seasonal adoption, has already been trained and validated on the multi-season SemanticBLT dataset.

\subsection{Ablation Study}
We conduct four ablation studies to: (i) determine the optimal penalties for incorrect hits and misses, (ii) evaluate the contribution of the noisy GPS prior to the likelihood function, (iii) assess the influence of each semantic class (trunks and poles) on the likelihood estimation, and (iv) examine the method’s generalisation to new environments.

\subsubsection{Penalty Scores}
\begin{figure}[t]
  \centering
  \includegraphics[width=0.98\linewidth]{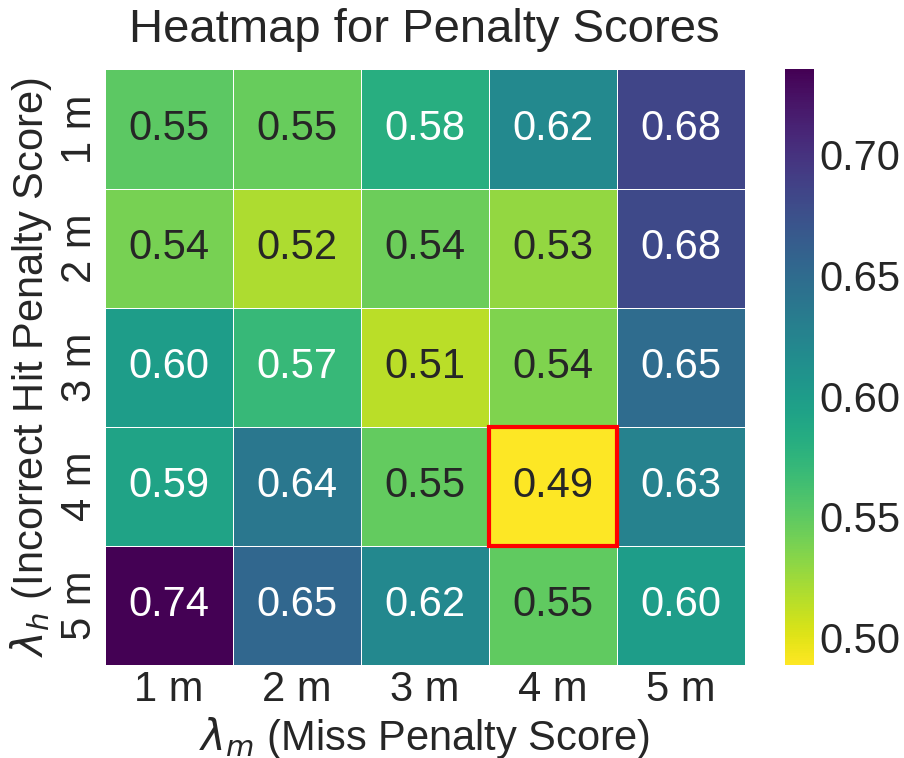}
  \caption{Heatmap of the average APE and RPE resulting from the ablation study on $\lambda_h$ and $\lambda_m$. The optimal configurations, yielding a minimum scores are highlighted by the red box.}
  \label{fig:heatmap}
\end{figure}

Optimal performance of the proposed particle filter is achieved by fine-tuning the parameters $\lambda_h$ and $\lambda_m$. To this end, the subsequent ablation studies analyse the effect of tuning each parameter individually. Fig. \ref{fig:heatmap} illustrates the heatmaps from the ablation study and the optimal parameter choices.

The penalty scores  $\lambda_h$ and $\lambda_m$ function similarly to an observation error in conventional particle filters. They were tuned over a range of 1~m to 5~m, corresponding to the maximum observation distance. We selected the optimal values by identifying the combination that yielded the lowest Absolute Pose Error (APE) and Relative Pose Error (RPE). Based on this ablation $\lambda_h = 4$~m and $\lambda_m = 4$~m were selected as the optimal values.

\begin{table}[t]
  \centering
  \caption{Comparison of relative pose error (RPE) and absolute pose error (APE) on semantic particle filter (SPF) and depth based particle filters (DPF). NGPS: Noisy GPS.}
  % Change the widths (in cm) to suit your layout needs
  \begin{tabular}{lcc}
    \toprule
    \multirow{2}{*}{\textbf{Ablation}} & \multicolumn{2}{c}{\textbf{Mean (Standard Deviation) [Unit: m]}}\\
    \cmidrule(lr){2-3}
    & APE & RPE \\
    \midrule
    1. Semantics Only & 3.62 (2.44) & 0.93 (0.77)\\
    2. Poles + NGPS & 1.27 (0.79) & 0.97 (0.75)\\
    3. Trunks + NGPS & 1.61 (1.16) & 0.97 (0.80)\\
    4. Generalisation & 1.50 (1.24) & 1.04 (0.76)\\ 
    \bottomrule
  \end{tabular}
  \label{tab:abl}
\end{table}

\subsubsection{Noisy GPS Contribution}
Equation~\ref{eq:lx} defines the overall likelihood as a combination of semantic and GPS terms. We removed the GPS component, reducing the formulation to $\log \mathcal{L}_\mathbf{x} = \log p_{\text{obs}}$. This experiment highlights the role of the noisy GPS prior in maintaining global localisation, particularly during row switching where semantic observations are sparse. As shown in the first row of Table~\ref{tab:abl}, the APE increases by a factor of three compared to the standard approach, underscoring the importance of incorporating noisy GPS into the proposed algorithm. Without this prior, the filter often drifts into incorrect rows and cannot reliably recover.

\subsubsection{Semantic Contribution}
To assess the contribution of each semantic class to the proposed particle filter, we conducted two experiments in which either trunks or poles were used in isolation. The second row of Table~\ref{tab:abl} reports the results when only poles are considered, while the third row reports results for trunks only. The findings show that removing poles increases the APE by 34\%, whereas excluding trunks leads to a modest increase of just 6\%. These results indicate that poles are considerably more informative than trunks for semantic particle filtering in vineyard environments.

\subsubsection{Generalisation}
To validate the robustness of the proposed particle filter, we conducted an additional trial in which the robot was driven through the same vineyard in the opposite direction at a different time of day from the dataset reported in Table~\ref{tab:evo}. The fourth row of Table~\ref{tab:abl} presents the APE and RPE values, which remain comparable to those observed in the original run, demonstrating the consistency of our approach. Importantly, the filter maintained localisation accuracy despite changes in viewpoint, illumination, and traversal direction, conditions under which geometry-based baselines typically degrade. This result highlights the invariance of semantic landmarks to environmental variation and confirms that the semantic particle filter generalises reliably across repeated trials in the same vineyard.

%%%%%%%%%%%%%%%%%%%%%%%%%%%%%%%%%%%%%%%%%%%%%%%%%%%%%%%%%%%%%%%%%%%%%%%%%%%%%%%%
\section{Conclusion}
\label{sec:conclusion}

In this paper, we presented a semantic particle filter for robust localisation in vineyards, addressing the limitations of geometry-based methods such as AMCL that suffer from row-level aliasing and drift. Our approach leverages panoptic segmentation to detect stable landmarks (trunks and poles), projects them into a BEV, and fuses them with LiDAR scans to form semantic observations for the particle filter. A central contribution is the introduction of semantic walls, which connect adjacent landmarks into pseudo-structural constraints that mitigate aliasing by enforcing row-level continuity. We further incorporate a dynamic GPS prior that preserves global consistency in headland regions where semantic cues are sparse.

Experiments in a real vineyard demonstrate that our method maintains localisation within the correct row, recovers from deviations where AMCL fails, and outperforms RTAB-Map by reducing overshoot and drift during row switching. These results highlight the value of exploiting semantic landmarks and pseudo-structural representations for localisation in highly repetitive agricultural environments, paving the way towards more reliable autonomy in field robotics.

Looking ahead, we plan to extend this framework to other structured agricultural domains such as orchards and forestry, where stable landmark distributions and pseudo-structural constraints could provide similar benefits for long-term and robust localisation.
%%%%%%%%%%%%%%%%%%%%%%%%%%%%%%%%%%%%%%%%%%%%%%%%%%%%%%%%%%%%%%%%%%%%%%%%%%%%%%%%
%% Future work: Use only if applicable -- but if so, use the following
%% sentence to start:
% Despite these encouraging results, there is further space for improvements. 
%% In general, I avoid explaining future work in 6-8 page conference papers...

%%%%%%%%%%%%%%%%%%%%%%%%%%%%%%%%%%%%%%%%%%%%%%%%%%%%%%%%%%%%%%%%%%%%%%%%%%%%%%%%
% Only if applicable
%\section*{Acknowledgments}
%We thank XXX for fruitful discussions and for \dots

\bibliographystyle{ieeetr}

% All new citations should go to new.bib. The file glorified.bib should go
% be the one from the ipb server. After paper or related work has been
% written merge the entries from new.bib to glorified.bib ON THE SERVER,
% replace the glorified.bib in this repository and empty the new.bib
\bibliography{glorified,new}

\end{document}